\documentclass[journal,comsoc]{IEEEtran}

\ifCLASSINFOpdf
\else
\fi

\usepackage{amsmath,amssymb,amsfonts,mathtools} 
\usepackage{booktabs}

\usepackage[LGRgreek]{mathastext}
\usepackage{graphicx}
\usepackage{subcaption}
\usepackage{multirow}
\usepackage{array}

\usepackage{hyperref}

\newcommand{\etal}{\textit{et al}. }
\newcommand{\ie}{\textit{i}.\textit{e}., }
\newcommand{\eg}{\textit{e}.\textit{g}., }

\begin{document}
\title{Occlusion-robust Visual Markerless Bone Tracking for Computer-Assisted Orthopaedic Surgery}


% \author{\IEEEauthorblockN{Xue Hu\IEEEauthorrefmark{1},
% Anh Nguyen\IEEEauthorrefmark{2}, and
% Ferdinando Rodriguez y Baena\IEEEauthorrefmark{1},~\IEEEmembership{Fellow,~IEEE}}
% \IEEEauthorblockA{\IEEEauthorrefmark{1}Mechatronics in Medicine Lab,
% Imperial College London, London, SW7 2AZ UK}
% \IEEEauthorblockA{\IEEEauthorrefmark{2}Hamlyn Centre, Imperial College London, London, SW7 2AZ UK}
% \thanks{Manuscript received December 1, 2012; revised August 26, 2015. 
% Corresponding author: Xue Hu (email: xue.hu17@imperial.ac.uk).}}

\author{Xue~Hu,
        Anh~Nguyen,
        and~Ferdinando~Rodriguez y Baena,~\IEEEmembership{Member,~IEEE}% <-this % stops a space
\thanks{Xue Hu and Ferdinando Rodriguez y Baena are with the Mechatronics in Medicine Lab,
Imperial College London, London, SW7 2AZ UK. e-mail: xue.hu17@imperial.ac.uk.}% <-this % stops a space
\thanks{Anh Nguyen is with the Hamlyn Centre, Imperial College London, London, SW7 2AZ UK.}% <-this % stops a space
% \thanks{Manuscript received April 19, 2005; revised August 26, 2015.}
}

% \markboth{IEEE Transactions on Instrumentation and Measurement,~Vol.~xx, No.~x, August~2015}%
\markboth{IEEE Transactions on xxx,~Vol.~xx, No.~x, August~2015}%
{Shell \MakeLowercase{\textit{et al.}}: IEEE Transactions on Magnetics Journals}

\IEEEtitleabstractindextext{%
\begin{abstract}
Conventional computer-assisted orthopaedic navigation systems rely on the tracking of dedicated optical markers for patient poses, which makes the surgical workflow more invasive, tedious, and expensive. Visual tracking has recently been proposed to measure the target anatomy in a markerless and effortless way, but the existing methods fail under real-world occlusion caused by intraoperative interventions. Furthermore, such methods are hardware-specific and not accurate enough for surgical applications. In this paper, we propose a RGB-D sensing-based markerless tracking method that is robust against occlusion. We design a new segmentation network that features dynamic region-of-interest prediction and robust 3D point cloud segmentation. As it is expensive to collect large-scale training data with occlusion instances, we also propose a new method to create synthetic RGB-D images for network training. Experimental results show that our proposed markerless tracking method outperforms recent state-of-the-art approaches by a large margin, especially when an occlusion exists. Furthermore, our method generalises well to new cameras and new target models, including a cadaver, without the need for network retraining. In practice, by using a high-quality commercial RGB-D camera, our proposed visual tracking method achieves an accuracy of $1-2^{\circ}$ and $2-4$ mm on a model knee, which meets the standard for clinical applications.
\end{abstract}

\begin{IEEEkeywords}
Biomedical applications, Image processing, Neural networks, Vision-based instrumentation and measurement.
\end{IEEEkeywords}}

\maketitle

\IEEEdisplaynontitleabstractindextext

\IEEEpeerreviewmaketitle

\section{Introduction}

\IEEEPARstart{R}{estoring} the mechanical axis of the lower limb is crucial in orthopaedic knee surgery \cite{mavrogenis2013computer}. For example, in total knee arthroplasty (TKA), the distal femur should be resected at a certain angle, and the prostheses should be congruently placed on the surrounding anatomy \cite{sugano2003computer}. However, up to 20\% of TKA procedures performed by experienced surgeons result in knee axis misalignment greater than 3$^{\circ}$ \cite{mahaluxmivala2001effect}. Implant misalignment could cause abnormal polyethene wear, joint instability and early implant failure, all of which would have a significant impact on patients' quality of life \cite{mavrogenis2013computer}. 

Over the past decade, navigation systems have been recognised as a powerful tool to improve the efficacy and accuracy of knee surgery \cite{siebert2002technique, vaccarella2013unscented}. By providing intraoperative measurements and pre-operative plannings in visual or numerical form, navigation systems guide the surgeon to reach the goal in various steps with greater control, precision, and consistency in time \cite{figueras2014surgical}. 
Conventional orthopaedic navigation systems usually embed an optical marker-based tracking mechanism to relate the computer-stored information to the actual patient pose on the surgical table. These systems therefore rely on a dedicated system to track the movement of passive or active infrared (IR) markers that are rigidly pinned and registered to the target bone. The patient-specific information could be image-based (\eg pre-operative or intra-operative medical scans such as CT or MRI) \cite{victor2004image} or image-free (\eg generic kinematic and/or morphological models parametrised onto the digitised anatomical landmarks) \cite{rebal2014imageless}. Surgeons need to first manually collect a set of key points such as implanted fiducials, anatomical landmarks or surface points, to which the image-based or image-free information can be registered by point-based \cite{hong2010effective} or surface-based approaches \cite{chen2014automated}. The registered initial target pose can then be updated according to the IR markers tracked by the optical tracker.

While marker-based tracking is currently regarded as the ``gold standard'' by many commercial navigation systems such as NAVIO (Smith \& Nephew PLC) and MAKO (Stryker Corp.), three main limitations exist: first, the marker incision causes an additional scar and further surgical exposure, which may increase the risk of infection, nerve injury, and bone fracture \cite{wysocki2008femoral,schulz2007results}. Second, surgeon involvement is required for marker preparation, fixation and marker-to-target registration. These steps have the potential to introduce additional human errors \cite{bae2011computer} and workload for surgeons \cite{beringer2007overview,pearle2010robot}. Finally, the bulky IR markers may interfere with surgeon's performance \cite{rodrigues2019deep}, as the immobile tracker requires constant line-of-sight to the target, which may restrict surgeon's movement in the operating room \cite{chan2017precision,vaccarella2013unscented}.

Thanks to the fast development in depth sensing, commercial RGB-D cameras can be explored to replace the dedicated optical system. Once the camera sees the exposed target, the pixels associated with the target are automatically segmented from the RGB-D frames by trained neural networks, then the segmented surface is registered to a reference model in real-time to obtain the target pose. Albeit the concise workflow \cite{hu2021markerless, yang2021automatic}, two aspects must be improved to move markerless tracking one step closer to surgical application \cite{hu2021head}: first, as both training data collection and network design consider no target occlusion, markerless tracking drifts during intraoperative interventions (\eg bone drilling). Therefore, the target must be kept still during manipulation, which is impossible for knee surgeries. Second, the networks are trained on a dataset collected by a single consumer-level RGB-D camera. Limited by the camera's quality, the achieved accuracy is below the clinical acceptance. A more precise camera is essential to achieve higher tracking accuracy. Ideally, the network should be adaptable to new cameras without retraining.

This paper presents a RGB-D markerless tracking method for knee surgeries, which is robust to target occlusions and better in precision. To do so, we propose a new deep neural network for automatic RGB-D segmentation and, since collecting and labelling a large number of training data are highly tedious and time-consuming, we augment the existing real data containing no occlusion instances with synthetic RGB-D data containing various simulated target interactions. We show that, by utilising both 2D RGB images and 3D point clouds converted from depth frames, our network successfully learns to be robust to occlusion from synthetic data only, and generalises well to new RGB-D cameras and knee targets. A video is provided in supporting file to demonstrate the success of our network in the real world.

Our contributions can be summarised as follows:
\begin{enumerate}
\item We propose a robust markerless bone tracking algorithm for orthopaedic navigation, which proves the usability of RGB-D tracking for surgical navigation.
\item We introduce a new large-scale synthetic RGB-D dataset generated with simulated target occlusion that allows network training in an effort-efficient way.
\item We conduct intensive experiments to verify the effectiveness of our network design under different cameras, lighting conditions, and synthetic-to-real transfer learning scenarios.
%\item Evaluation of tracking performance on different RGB-D cameras and/or targets.
\item Our method achieves clinically acceptable tracking error on a model leg with a high-quality commercial RGB-D camera. To the best of our knowledge, this is the first study that verifies the suitability of visual markerless tracking for clinical applications. 
% \textbf{This is a bold claim. Make sure we have enough exp. to support this.}
\end{enumerate}

The rest of the paper is organised as follows. Starting with related work in Section II, we describe our methodology in Section III. Section IV presents an evaluation of network accuracy on real test data collected under occlusion. Section V shows the performance of markerless tracking on different targets and for different RGB-D cameras. Finally, we conclude the paper and discuss the future work in Section VI.

\section{Related Works}
\subsection{Pose Measurement for Surgical Guidance}
Research effort has been dedicated to improving the robustness and cost effectiveness of current navigation systems. Some studies combined the optical tracking with additional measurements to solve the line-of-sight problem: Vaccarella \etal fused optical and electromagnetic tracking based on an unscented Kalman filter \cite{vaccarella2013unscented}; Enayati \etal synchronised optical and inertial data by a quaternion-based unscented Kalman Filter \cite{enayati2015quaternion}; Ottacher \etal proposed a compact 3D ultrasound system that combines conventional 2-D ultrasound with optical tracking \cite{ottacher2020positional}. Alternatively, for surgeries such as maxillofacial surgery \cite{suenaga2015vision} whose target is relatively clean and feature-rich, prominent anatomy features can be detected from RGB recording and registered for poses \cite{liu2017new}. For surgeries with complex scenes (e.g., where the target is surrounded by blood and tissues), depth-sensing is exploited to improve the detection robustness. For example, Sta \etal proposed a point‐pair features algorithm to estimate the pose of TKA implant from depth captures, but such a feature-based method cannot be run in real-time.

% \subsection{Deep Learning with Synthetic Data}
% Simulated RGB-D data have been  for network training and how to overcome the simulation-to-reality gap is the key to successful training. In \cite{james2017transferring}, an end-to-end reactive controller was trained on simulated RGB videos and planned trajectories to control the robot velocities for a multi-stage task. The authors suggested that texture randomisation helps to generalise the learned RGB knowledge to reality. In \cite{shrivastava2017learning}, a large number of RGB eye images were generated by a simulator with ground truth gaze directions to train a gaze predictor. The synthetic images were enhanced by an unsupervised Generative Adversarial Network (GAN) to reduce the domain gap.  While unsupervised techniques (\eg GAN) can still be applied \cite{baruhov2020unsupervised}, supervised networks can generate more realistic noise profiles for over-simplistic synthetic depth maps \cite{sweeney2019supervised}. Alternatively, in the work by \cite{mueller2017real}, half-synthetic RGB-D data were created to train a hand pose predictor, by merging the real background captures with a sampled virtual hand controlled in a mixed reality approach \cite{mueller2017real}. 

\subsection{RGB-D Learning-based Markerless Tracking}
Commercial RGB-D cameras can achieve fast and
accurate measurement in a high resolution, making them potential new tools for surgical navigation. For real-time tracking purposes, learning-based methods were explored in the literature to extract comprehensive features automatically. Yang \etal designed a fully convolutional network (FCN) to automatically segment spine area from RGB-D captures so that the pre-plannings can be overlayed accordingly \cite{yang2021automatic}. The RGB and depth features were encoded from input 2D maps, fused at different stages of encoder, and decoded jointly to predict the segmentation mask \cite{yang2021automatic}. Liu \etal proposed a sequential RGB-D network for automatic femur tracking \cite{heliu}. The target centre was roughly localised by a convolutional neural network (CNN) in the RGB frame first. Then the aligned depth map was cropped around the predicted centre with a fixed size of 160$\times$160 pixels, and passed to another CNN to finely predict the femur mask. The femur area was segmented from depth maps according to the prediction, converted to 3D points, and registered to a scanned model by iterative closest point (ICP) in real-time. 
% \textbf{Careful with this claim, depth image has some problems as well, \eg distance, material, and also lighting (depend on the tech used in the camera)}

Unlike these literature methods which focus on a clean target surface and train the network with collected real data \cite{yang2021automatic, heliu}, we aim to improve the tracking robustness when the target is manipulated under occlusion, by generating a synthetic dataset to train a segmentation network with new design. 

\section{Synthetic Data Creation}

While the available dataset collected in \cite{heliu} by a RealSense D415 camera (Intel Corp.) has a limited size (5200 independent frames on a cadaver knee and a model knee) and contains no target occlusion, a large dataset with occlusion instances is essential to train a network that works within an intraoperative scenario. To expand the current training data in a fast and efficient way, we generate synthetic data using a modular procedural pipeline, BlenderProc \cite{denninger2019blenderproc}, on an open-source rendering platform, Blender \cite{blender}. 
% Compared to OpenGL rendering that relies on simple rasterisation, Blender rendering is more photorealistic regarding the shading consistency, physically constrained sampling, and object interaction in proximity \cite{denninger2019blenderproc}. 
The details of data generation are described below.

\subsubsection{Creation of Randomised Scenes} A model knee is scanned by a highly precise scanner (HDI 3D scanner, LMI Technologies Inc.). The obtained frames are co-registered into a single point cloud. After hole filling \cite{liepa2003filling} and Laplacian normal smoothing \cite{herrmann1976laplacian}, a 3D knee model is reconstructed from the merged point cloud by application of the Screened Poisson algorithm \cite{kazhdan2013screened}. Then, the model is manually divided into the femur and not-femur sections and imported into the Blender pipeline via the Python scripting interface.

As suggested in \cite{james2017transferring}, domain randomisation is critical to overcoming the simulation-to-reality gap in RGB data. Therefore, we randomly alter the scene during image generation regarding (\autoref{fig:examples}):
\begin{itemize}
\item The type (point or surface) and strength of lighting.
\item The room background, which contains arbitrary extrusions and objects loaded from the Ikea dataset \cite{lpt2013ikea} as distractors. The materials of the wall, floor and loaded objects are randomly sampled from a large public material database, ambientCG \cite{ambientcg}.
\item The material of skin and exposed bone, by blending a random texture with a random RGB colour.  
\end{itemize}

\subsubsection{Simulation of Foreground Occlusion}
Five 3D models of human hands and surgical tools are prepared and imported as foreground distractors. These objects are randomly positioned and orientated within the camera's line-of-sight of the exposed femur to simulate partial target occlusion. The fingertip or tooltip, defined as the origin of local object coordinates, can optionally contact and translate on the femur surface. The material of these objects is also altered following the random texture blending method mentioned above.  

\subsubsection{Adding Depth Sampling Noise}
The simulation-to-reality gap was found to be more significant in depth imaging due to the complex noise profiles and diverse ecosystem of sensors \cite{sweeney2019supervised}.
A commercial structured-light depth camera mainly experiences three kinds of sampling noise:  
\begin{itemize}
\item Pixel location offsets due to quantised disparity \cite{handa2014benchmark}: the final depth value at a pixel $Z(x, y)$ is interpolated from the raw sampling at adjacent pixels $Z(x+\delta _x, y+\delta_y)$.
\item The IID Gaussian deviation of depth values, presumably due to sensor noise or errors in the stereo algorithm \cite{barron2013intrinsic} (\ie $Z(x, y)=\hat Z(x, y)+\delta _z$). 
\item The depth value dropout at some pixels (\ie $Z(x_d, y_d)=0$) due to two reasons: the interaction of the projected IR pattern with illumination and target material, or the interaction of the depth camera’s projector-receiver pair with the scene \cite{sweeney2019supervised}. 
% Pixels around highly angled surfaces or boundaries are especially vulnerable to such artefacts.  
\end{itemize}
Modelling the dropout noise is extremely challenging, since the material- and illumination-dependent interaction can not be physically simulated. Besides, the dropout density is subject to specific camera properties like spatial sampling resolution and the baseline distance between projector and receiver. Therefore, we model the first two types of noise in a Gaussian distribution with $(\delta_x, \delta_y)\sim \mathcal{N}(0, 1/2)$ and $\delta _z\sim0.08\mathcal{N}(0, 1/3)$ (in mm), according to the datasheet of D415 camera \cite{grunnet2018best}. 

\begin{figure}[htbp]
\centering
\includegraphics[width =0.5\textwidth]{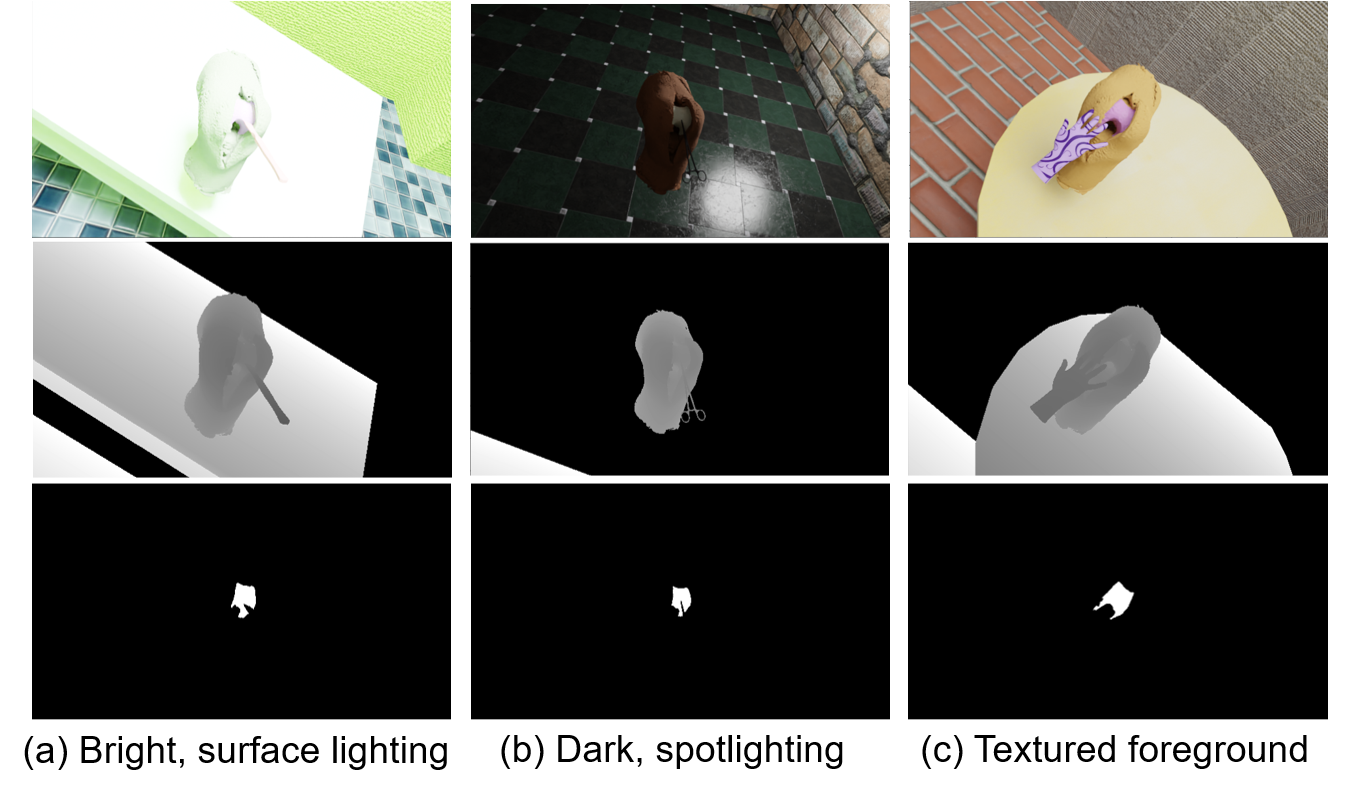}
\caption{Example synthetic images with variations in the strengths and types of lighting, background and foreground.}
\label{fig:examples}
\end{figure}

\begin{figure*}[htbp]
\centering
\includegraphics[width =\textwidth]{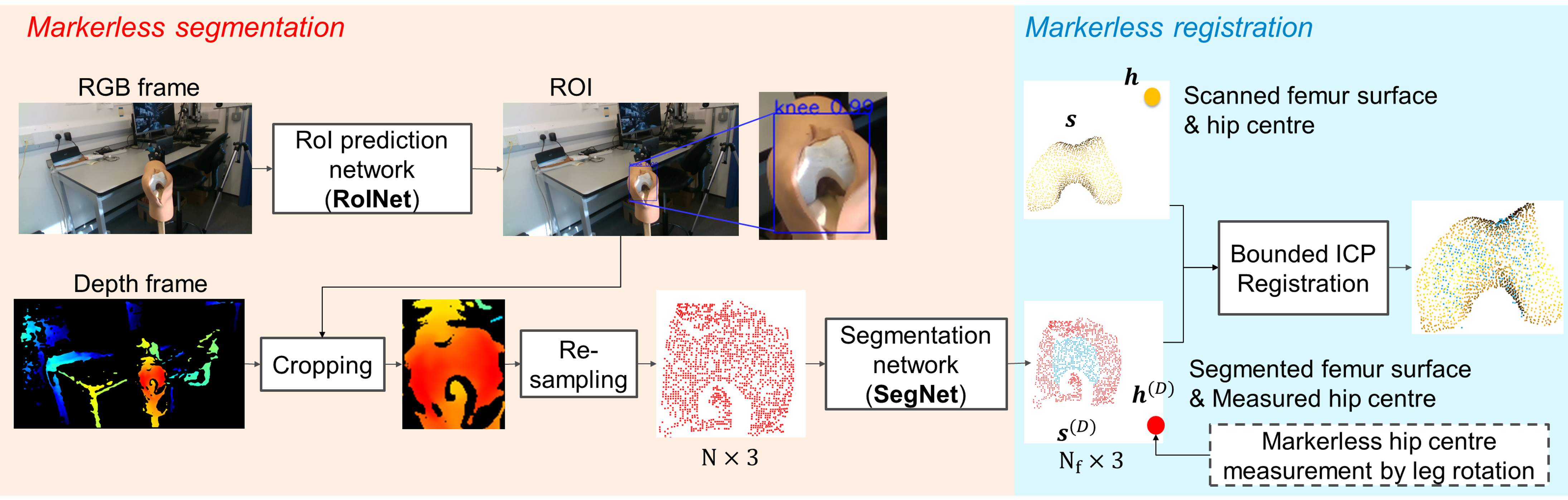}
\caption{Overview of markerless tracking: Starting from an aligned RGB-D frame, we initially compute the RoI for the exposed target femur using our RGB-based RoINet. After cropping the depth frame with predicted RoI, a N$\times3$ point cloud is resampled and input to the segmentation network to predict the femur label for every point. The $N_f$ segmented femur points are then registered to a pre-scanned reference model by a Bounded ICP algorithm implementation in real-time to obtain the target pose. }
\label{fig:overview}
\end{figure*}

\subsubsection{Statistics}
For each image-generation session with a settled scene, 20 captures are taken with random camera poses. The viewpoint is controlled to be 0.5-1m away from the target to replicate the physical working distance. The sampling intrinsic parameters and resolution are set to the physical values of a RealSense camera calibrated by a standard routine \cite{Zhang2000:CameraCalib}. The visibility of the exposed femur is checked for each sampling pose to ensure a meaningful capture. The simulation is repeated to produce $10,000$ randomised synthetic RGB-D images together with automatically labelled binary femur masks. \autoref{fig:examples} shows some examples of generated synthetic images.

\section{Markerless Segmentation and Registration}
\autoref{fig:overview} shows an overview of the proposed markerless tracking workflow. The whole procedure can be divided into two steps: automatic target segmentation and real-time pose registration. In this section, we will describe how we implement each part for better tracking robustness and accuracy.

\subsection{Automatic Segmentation Network}
Similar to \cite{heliu}, our segmentation network contains a sequential arrangement to leverage both RGB and depth imaging (\autoref{fig:overview}). The stable RGB stream ensures robust target localisation in the full scope of captures, while the depth data ensure fine segmentation, as they are less impacted by bleeding and surgical lighting \cite{sta2021towards}. The RoI box is first predicted from the global RGB frame by a RoINet, according to which the aligned depth frame is cropped and resampled into a 3D point cloud. A SegNet then predicts the femur mask from the cloud for point-wise segmentation. The details for both networks are explained below.

\begin{figure*}[htbp]
\centering
\includegraphics[width = 0.9\textwidth]{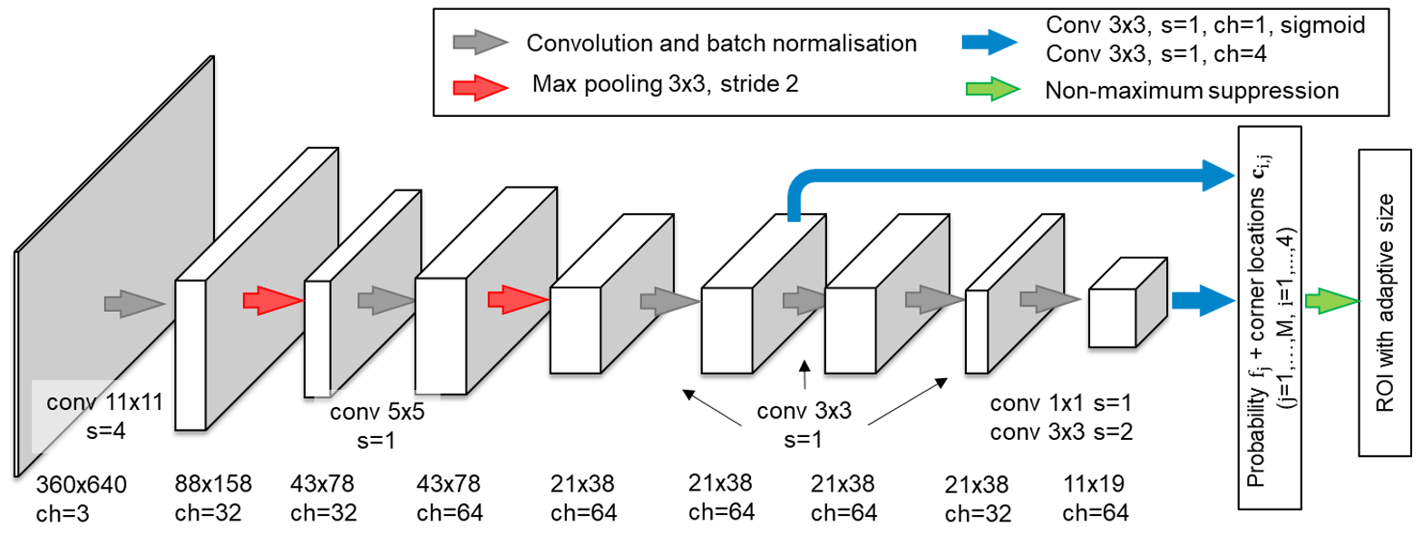}
\caption{Architecture of the RoI prediction network based on RGB information. We extract feature maps by an Alexnet backbone, and take multi-scale features for multi-box classification and corner regression. In our implementation, M=21$\times$38+11$\times$19.}
\label{fig:roinet}
\end{figure*}
\subsubsection{RoINet}
Unlike \cite{heliu}, where the authors regress a target centre location and crop the depth frames with a fixed box size to match the input dimension of the segmentation CNN, our network directly predicts an RoI box with an adaptive size to more tightly bound the exposed femur surface. The change is required for two reasons: first, to ensure high segmentation speed for real-time tracking, only a certain number of resampled points (N) could be taken by the segmentation network. However, a sufficient number of segmentation outputs ($N_f$) are desired for reliable pose registration. Therefore, RoI cropping should ensure a high target occupation rate $N_f/N$. Second, when the camera moves towards or away from the target, or the network is deployed to a new camera with a considerably different focal length, a fixed cropping size may fail to cover the whole target dimensions. Therefore,  RoI cropping should be dynamic in size to ensure a nearly constant value for $N_f/N$.

The RoINet, as shown in \autoref{fig:roinet}, is modified from the localisation network proposed in \cite{heliu}, by adding two mid-layer auxiliaries and a multi-box loss function. With a similar design to Alexnet, the first five convolutional layers extract feature maps with shrinking sizes from the input RGB image. Inspired by the Single Shot Multibox Detector (SSD) \cite{liu2016ssd}, M multi-scale feature maps are taken from different layers and convoluted by 3$\times$3 kernels to produce M bounding boxes with a probability for the presence of the target in the box (0\textless$f$\textless1)). Each bounding box $\mathbf{c}=[c_1, c_2, c_3, c_4]$ is uniquely decided by the x and y offset of upper-left and lower-right corners relative to the default box coordinates of [-0.5, -0.5, 0.5, 0.5]. The overall $M\times(4+1)$ predictions are processed by a non-maximum suppression to decide the best RoI box.

\begin{figure*}[htbp]
\centering
\includegraphics[width =1\textwidth]{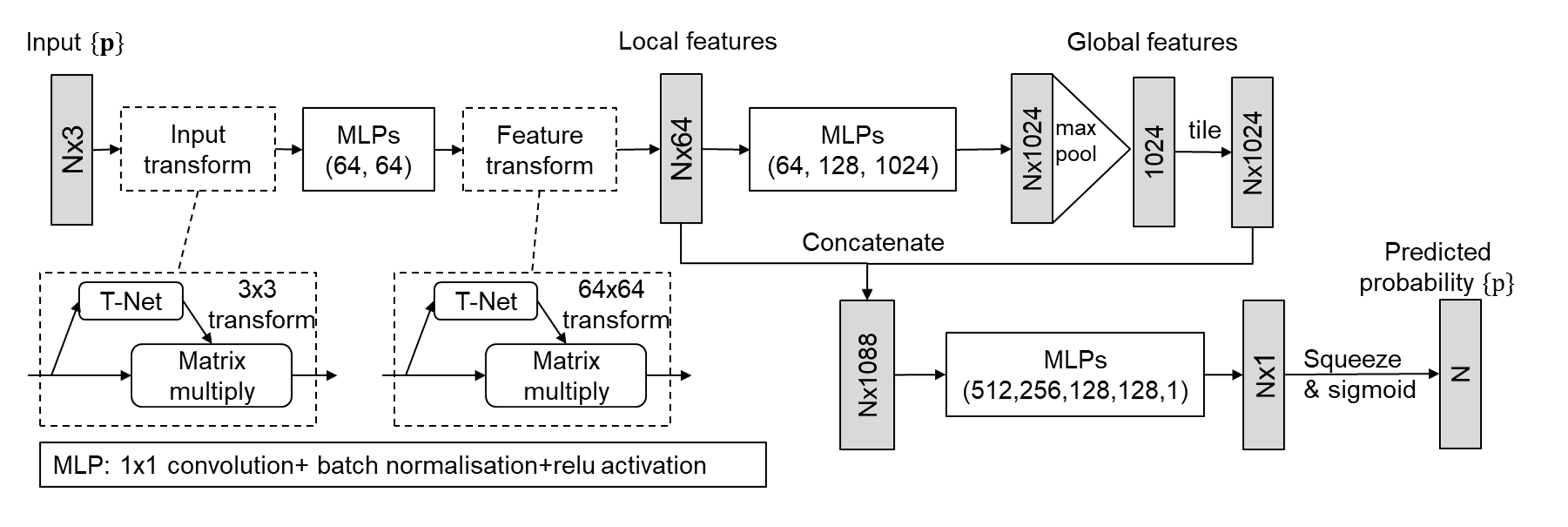}
\caption{Architecture of the 3D SegNet. N$\times$3 input points are resampled from the cropped depth frames according to the predicted RoI. The encoder-decoder structure follows the design of PointNet. The decoded 1-channel output is additionally processed by a sigmoid function to predict the probability. N=2000 in our implementation.}
\label{fig:segnet}
\end{figure*}
\subsubsection{SegNet}
The generated depth maps shown in \autoref{fig:examples} apparently lack realistic depth dropout. Fortunately, compared to 2D depth maps, the 3D point cloud representation of depth data is less vulnerable to such sampling artefacts (Section \ref{sec:ablation}).
Network-learned features should be similar in both real and synthetic domains to ensure knowledge transfer; they should also be robust to camera sampling properties so that the trained network is camera-agnostic.

Consequently, as shown in \autoref{fig:segnet}, our SegNet is designed to learn from the 3D point cloud representation rather than the 2D depth maps (as used in \cite{yang2021automatic, heliu}). It takes over the PointNet architecture \cite{qi2017pointnet} to predict from an $N\times3$ input point cloud $\{\mathbf{p}\}$. 
The input points are processed by a succession of Multi-Layer Perceptrons (MLPs) to produce $N\times 1024$ encoded features. A symmetric maximum pooling function is applied to extract a 1024-dimensional global descriptor, which is then concatenated with a local feature vector taken from a mid-layer. Next, the combined latent features are decoded by MLPs and reshaped into an N-dimensional vector. The predicted vector is finally mapped between 0-1 by a sigmoid function. The output values $\{p\}$ represent how possible is it that each of the N points belongs to the femur surface. The point with predicted probability $p_j$ higher than a threshold (0.8 in our implementation) can be regarded as a target femur point.

\subsubsection{Network Training}
The whole dataset is randomly divided into training and validation sets by a ratio of 8:2. The two networks are separately trained using the Tensorflow library \cite{tensorflow}. For the RoINet, the batch size is set to 4 and the training loss $\mathcal{C}_{\scriptscriptstyle{RoI}}$ is defined as the total difference of predicted box corner locations and probabilities ($c_{i,j}, f_j$) of the labelled values ($\hat{c}_{i,j}, \hat{f}_j$):
\begin{equation}
  \mathcal{C}_{\scriptscriptstyle{RoI}} = \sum\limits_{j=1}^{M}(\sum\limits_{i=1}^{4}|c_{i}-\hat{c}_{i}|_j+|f_j-\hat{f}_j|)
\end{equation}
For the SegNet, the batch size is set to 32 and the training loss $\mathcal{C}_{\scriptscriptstyle{Seg}}$ is defined as the sum of absolute differences between the predicted probabilities $p_j$ and binary femur labels $\hat{p}_j$.
\begin{equation}
\mathcal{C}_{\scriptscriptstyle{Seg}} = \sum\limits_{j=1}^{N}|p_j-\hat{p}_j|
\end{equation}
The Adam optimiser with an exponentially decaying learning rate starting from 0.001 is used for both training to ensure a steady rate of learning.
% Both networks were trained from scratch until convergence with no further reduction in validation loss. The training time is approximately 3 days on an NVIDIA Titan X GPU 

\subsection{Markerless Registration}
% The segmentation network automatically identifies a clean femur surface $\mathbf{s}^{\scriptscriptstyle{(D)}}$ from RGB-D captures. A pre-scanned reference model can then be registered in real-time to the acquired surface for target pose $\mathbf{P}^{\scriptscriptstyle{(D)}}(t)$ (\autoref{fig:overview}). 
As the segmented points have a limited spatial spread over the partially exposed femur area, classical ICP-based registration is vulnerable to rotational misalignment \cite{rodriguez2013bounded}. For higher registration accuracy and better robustness against wrongly segmented points, we thus adopt a previously validated and published Bounded ICP (BICP) method \cite{ hu2021markerless}. 
% The alignment of femoral mechanical axis, especially in terms of the varus–valgus and anterior–posterior orientation, can be effectively bound by the intra-operatively estimated hip centre
BICP uses a remote pair of corresponding features (\eg the model hip $\mathbf{h}$ and measured hip centre $\mathbf{h}^{\scriptscriptstyle{(D)}}$) to bound the registration error between the scanned model surface $\mathbf{s}$ and the automatically segmented femur surface $\mathbf{s}^{\scriptscriptstyle{(D)}}$:
\begin{equation}
    \mathbf{P}^{\scriptscriptstyle{(D)}}(t)
    = BICP(\mathbf{s}^{\scriptscriptstyle{(D)}}(t), \mathbf{s}, \mathbf{h}^{\scriptscriptstyle{(D)}}(t), \mathbf{h})
\end{equation}
$\mathbf{s}$, $\mathbf{h}$  and $\mathbf{h}^{\scriptscriptstyle{(D)}}(t)$ are obtained with a setup similar to that proposed in \cite{hu2021markerless}:

\subsubsection{Model surface and hip location} 
\begin{figure}[htbp]
\centering
\includegraphics[width =0.45\textwidth]{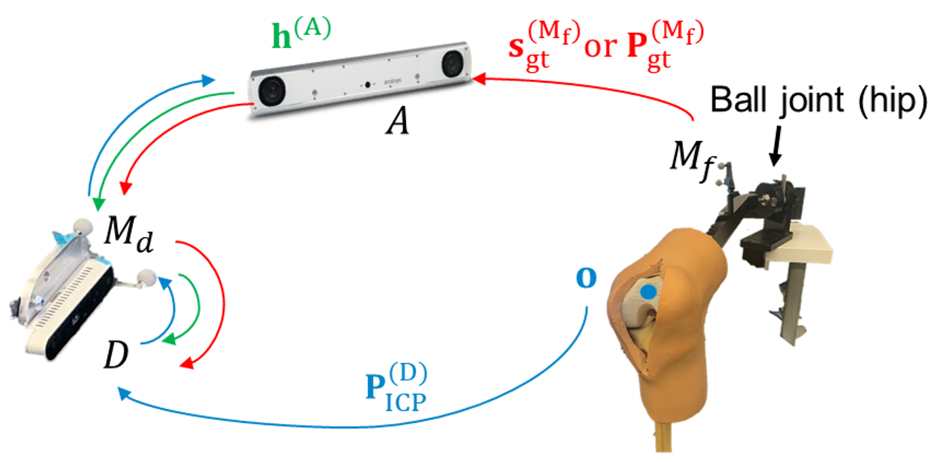}
\caption{The system setup for BICP registration and evaluation. Blue lines: transformation from the tracked landmark $\mathbf{o}$ into $\mathbf{o}^{\scriptscriptstyle{(A)}}$ for hip centre calculation; Green lines: online transformation of the fitted hip centre $\mathbf{h}^{\scriptscriptstyle {(A)}}$ into $\mathbf{h}^{\scriptscriptstyle {(D)}}$. Red lines: transformation of ground truth femur surface or pose for labelling or evaluation.}
\label{fig:hip}
\end{figure}

In our laboratory setup, before online tracking, the model surface $\mathbf{s}$ is digitised from the femur under maximum skin exposure. The hip centre $\mathbf{h}$ is sphere-fitted from the probed surface points of the ball joint (\autoref{fig:hip}). In a clinical setup, the model $\mathbf{s}$ and $\mathbf{h}$ would instead be reconstructed from pre-operative images such as CT or MRI. 

\subsubsection{Measured hip location}
As shown in \autoref{fig:hip}, an optical marker $\mathit{M_{\scriptscriptstyle{d}}}$ is rigidly anchored to the depth camera $D$, which is tracked by a global optical tracker $\mathit{A}$ (FusionTrack 500, Atracsys LLC.) to obtain a hip centre estimate. The hip centre can be  modelled as the pivot point around which the leg is rotated. To track a femur landmark during rotation in a markerless way, we combine the aforementioned automatic segmentation with ICP registration to track a rough femur pose $\mathbf{P}^{\scriptscriptstyle{(D)}}_{\scriptscriptstyle{ICP}}(t)$. The local model origin $\mathbf{o} = [0,0,0,1]^{\scriptscriptstyle T}$ is chosen as the landmark to avoid any projection bias due to the rotational registration error. The landmark positions tracked by ICP-based markerless tracking are transformed into global coordinates by the hand-eye calibrated transformation $\prescript{\scriptscriptstyle{M_{\scriptscriptstyle{d}}}}{\scriptscriptstyle D}{\mathbf{T}}$ and optically tracked $\prescript{\scriptscriptstyle A}{\scriptscriptstyle{M_{\scriptscriptstyle{d}}}}{\mathbf{T}}$ as follows:
\begin{equation}
    \mathbf{o}^{\scriptscriptstyle{(A)}}(t)
    = \prescript{\scriptscriptstyle A}{\scriptscriptstyle{M_{\scriptscriptstyle{d}}}}{\mathbf{T}}(t)  \times
    \prescript{\scriptscriptstyle{M_{\scriptscriptstyle{d}}}}{\scriptscriptstyle D}{\mathbf{T}} \times
    \mathbf{P}^{\scriptscriptstyle{(D)}}_{\scriptscriptstyle{ICP}}(t) \times \mathbf{o}
\end{equation}
During rotation, more than 40 frames of $\mathbf{o}^{\scriptscriptstyle{(A)}}$ are recorded, from which the still hip centre $\mathbf{h}^{\scriptscriptstyle{(A)}}$ is computed by a sphere-fitting algorithm \cite{hu2021markerless}. The estimated global hip location $\mathbf{h}^{\scriptscriptstyle{(A)}}$ is finally transformed back to the depth camera frame as $\mathbf{h}^{\scriptscriptstyle {(D)}}(t)$ by $\prescript{\scriptscriptstyle {M_{\scriptscriptstyle{d}}}}{\scriptscriptstyle{A}}{\mathbf{T}}(t)$ for online BICP registration (green path in \autoref{fig:hip}).

\section{Network Evaluation}

\subsection{Test Data Collection}
To evaluate the performance of the trained segmentation network in the real world, we collected 800 RGB-D captures by a RealSense D415 camera, during which the target femur was partially occluded by hands or tools. To automatically label the femur pixels, an optical marker $\mathit{M_{\scriptscriptstyle{f}}}$ was inserted into the metal leg so that the ground truth (gt) femur surface could be optically tracked (red path in \autoref{fig:hip}). 
After a standard exposure, the femur surface was manually digitised as $\mathbf{s}_{\scriptscriptstyle{gt}}^{\scriptscriptstyle{(A)}}$. The probed surface was then calibrated to $\mathit{M_{\scriptscriptstyle{f}}}$ as $\mathbf{s}_{\scriptscriptstyle{gt}}^{\scriptscriptstyle{(M_{\scriptscriptstyle{f}})}}$, and further transformed into $D$ according to:
\begin{equation}
    \mathbf{s}_{\scriptscriptstyle{gt}}^{\scriptscriptstyle{(D)}}(t)
    =  \prescript{\scriptscriptstyle D}{\scriptscriptstyle{M_{\scriptscriptstyle{d}}}}{\mathbf{T}}
    \times \prescript{\scriptscriptstyle {M_{\scriptscriptstyle{d}}}}{\scriptscriptstyle{A}}{\mathbf{T}}(t)
    \times \prescript{\scriptscriptstyle A}{\scriptscriptstyle{M_{\scriptscriptstyle{f}}}}{\mathbf{T}}(t) 
    \times \mathbf{s}_{\scriptscriptstyle{gt}}^{\scriptscriptstyle{(M_{\scriptscriptstyle{f}})}}
\end{equation}

\begin{figure}[htbp]
\centering
\includegraphics[width =0.48\textwidth]{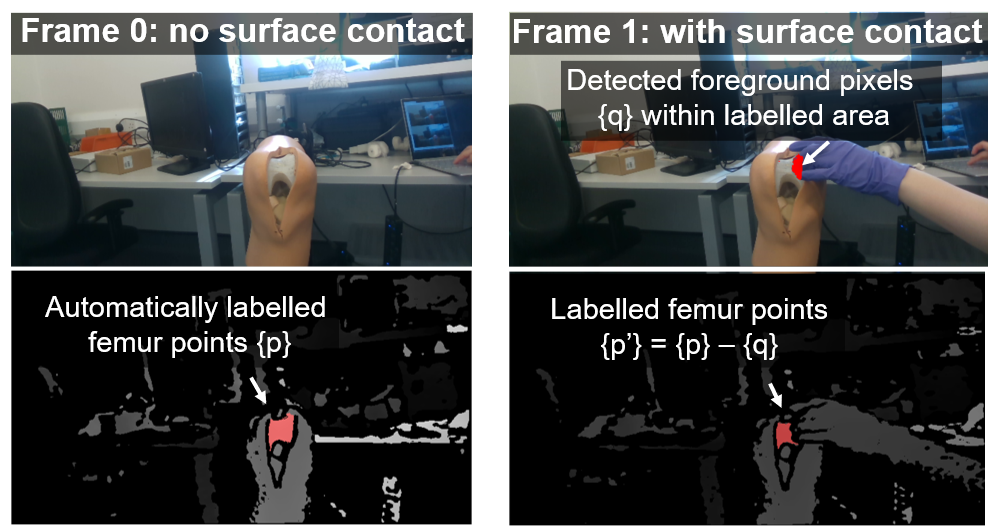}
\caption{Generation of the ground truth label mask for a target femur under surface contact based on a pairwise capture.}
\label{fig:label}
\end{figure}

As suggested by \cite{heliu}, the transformed surface points $\mathbf{s}_{\scriptscriptstyle{gt}}^{\scriptscriptstyle{(D)}}$ were finally registered to the raw depth capture by a standard ICP algorithm to identify the matching pixels that should be labelled as femur points. However, when hands or tools occluded the target surface, the registration between digitised surfaces and unsegmented captures became highly unreliable. To ensure correct annotation under target occlusion, we utilised pairwise captures. As shown in \autoref{fig:label}, the target was first captured with no surface occlusion or contact, then labelled by ICP-based point matching as described above (frame 1). Subsequently, without moving the camera or target, another capture was carried out for the femur surface while being partially occluded by a hand in a purple glove or a tool wrapped in purple tape to simplify the segmentation process, as follows. The femur mask labelled in frame 1 was applied to frame 2's RGB frame to segment an RoI, which was then converted to hue saturation and value (HSV) format, and filtered by a band-pass hue filter in the purple colour range to identify the pixels that belong to the foreground. The gt femur pixels for frame 2 were finally computed by subtracting the femur pixels in frame 1 by the detected foreground pixels in frame 2. The gt RoI box was computed as the smallest rectangle that covers all gt femur pixels.  

\subsection{Results}

\begin{figure*}[htbp]
\centering
\includegraphics[width =\textwidth]{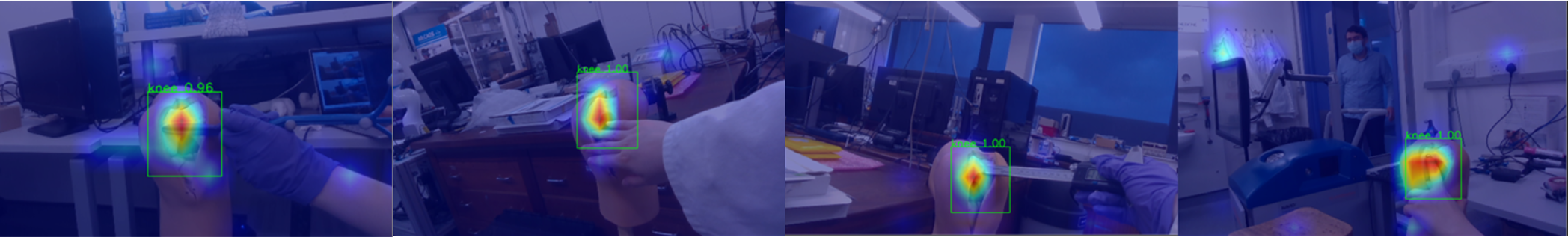}
\caption{The femur class GRAD-CAM activation heat map with the predicted ROI box and confidence.}
\label{fig:gradcam}
\end{figure*}
\autoref{fig:gradcam} shows some example images with overlaid Grad-CAM heat maps obtained by the proposed RoI prediction network. Regardless of hand occlusion, tool manipulation, capturing perspective and human presentation, the network properly pays attention to the exposed femur.  
If the intersection over union (IoU) between the predicted RoI and gt RoI is higher than 0.5, the prediction is regarded as successful. The overall accuracy is presented by the success rate of predictions over the entire test dataset. The localisation network trained in \cite{heliu} is also tested as a reference for comparison. The predicted RoI is regarded as the box drawn around the inferred target location, with the same size as the ground truth RoI box. 

Depending on the gt label (positive: is femur; otherwise negative) and the correctness of prediction (true: prediction matches gt; otherwise false), the N points can be classified as true positive (TP), true-negative (TN), false positive (FP) and false-negative (FN). To avoid the bias arising from a large number of TN predictions for background points, the segmentation accuracy is defined as the IoU score in each frame:
\begin{equation}
     IoU = \frac{TP}{TP+FP+FN}
\end{equation}
The overall accuracy is presented by the mean and standard deviation of IoU values over the full dataset. \autoref{tab:result} lists the evaluated accuracy of our networks and the reference networks proposed in \cite{heliu}. Our networks are almost twice more accurate than the reference networks.

\begin{table}[htb]
\centering
\caption{Accuracy comparison of RoI prediction and point/pixel segmentation, between our networks and the reference networks proposed in \cite{heliu}.}
\label{tab:result}
\begin{tabular}{l|l|l|l}
\bottomrule
                 & RoI    & Seg       & RoI+Seg   \\ \bottomrule
Liu \etal \cite{heliu} & 67.54\% & 42.03$\pm$32.96\% & 39.45$\pm$30.18\% \\\hline
\textbf{Ours} & 94.78\% & 85.42$\pm$12.43\% & 84.20$\pm$14.43\%   \\ \toprule
\end{tabular}
\end{table}

\subsection{Ablation Study}\label{sec:ablation}
The higher accuracy of our trained networks may be due to the new network structures, or the synthetic data included for training. We run ablation tests to study the effect of each component. Specifically, we want to answer three questions:
\begin{enumerate}
\item Are the synthetic data helpful in improving the robustness to occlusion for our networks?
\item Can other (\eg Liu \etal \cite{heliu}) networks be improved by learning on synthetic data? \item What is the critical factor that causes a difference in transferring ability?
\end{enumerate}

% \textbf{The next few paragraphs are quite difficult to understand. We should somehow explain or organize Table II in a better way. (There are too many "As shown in Table II", - things are not clear)}
To answer the first two questions, we additionally trained the proposed networks on the real part of the data only, and the reference network \cite{heliu} on our synthetic-included dataset. By comparing the ``Real'' with ``Real+sim'' group shown in \autoref{tab:ablation}, the simulated images significantly improve the robustness of our RGB and depth networks against real-world occlusion, while it harms the reference networks \cite{heliu}. 

\begin{table}[htb]
\centering
\caption{Ablation study for the effect of synthetic images and network structure.}
\label{tab:ablation}
\begin{tabular}{p{1.5cm}|l|l|l}
\toprule
Network                   & Training data               & RGB & D \\ \bottomrule 
\multirow{3}{*}{Liu \etal \cite{heliu}} & Real                        & 67.54\%   &  42.03$\pm$32.96\% \\ \cline{2-4} 
                        %   & Real+sim                    & 60\%    &  47.81$\pm$31.22\% \\ \cline{2-4} 
                          & Real+sim &  60\%    &  0 \\ \cline{2-4} 
                          & \begin{tabular}[c]{@{}l@{}}Real+sim \\ with dropout\end{tabular}                   & -   &  57.81$\pm$31.22\% \\ \bottomrule 
\multirow{3}{*}{\textbf{Ours}} & Real                        & 52.87\%    & 76.81$\pm$17.83\%  \\ \cline{2-4} 
                          & Real+sim                    & 94.78\%    & 85.42$\pm$12.43\%   \\ \cline{2-4} 
                          & \begin{tabular}[c]{@{}l@{}}Real+sim \\ with dropout\end{tabular} & -    &  85.37$\pm$11.70\%  \\ \bottomrule
\multirow{2}{*}{\begin{tabular}[c]{@{}l@{}}Ours without\\ RGB auxiliary\end{tabular}} & Real                                                                       & 32.95\%    &  - \\ \cline{2-4} 
                                        &     Real+sim                                                                    &  74.80\%   & -  \\ \toprule
\end{tabular}
\end{table}

For the last question, we investigated the segmentation network first. The proposed structure learns 3D geometric features from an unorganised point cloud, whereas the reference structure learns 2D features from a cropped depth map. As shown in \autoref{fig:dropout}, the depth dropout artefact makes the simulated 2D depth maps clearly different from real captures, but has less effect on the converted 3D point cloud since both data are sampled from the same 3D geometry. To prove the correlation between depth dropout and transferring ability, we generated 5,000 synthetic depth frames with simulated partial dropout noise caused by the interaction between scene and projector-receiver: an extra viewpoint was set up in Blender as the pattern projector in addition to the main viewpoint as the signal receiver. The ray cast from the projector to each sampled pixel was computed to find the pixels that cannot receive projected patterns. The depth values of those pixels were then overridden by zeros, resulting in a more realistic 2D depth map (\autoref{fig:dropout}). Our proposed network and reference segmentation network \cite{heliu} were then trained with dropout-included synthetic data. By comparing the ``Real+sim'' and ``Real+sim with dropout'' group shown in \autoref{tab:ablation}, as expected, the partially modelled dropout artefact improves the knowledge transfer for the reference network, but makes no difference to the proposed network. 

\begin{table}[h!]
  \centering
  \begin{tabular}{wc{0.5cm} | wc{2cm} | wc{2cm} | wc{2cm} }
    \bottomrule
    & Real capture & \begin{tabular}[m]{@{}l@{}}Simulated, no\\ dropout noise\end{tabular} 
 & \begin{tabular}[m]{@{}l@{}}Simulated, with\\ dropout noise\end{tabular} \\ \bottomrule
   \begin{tabular}[m]{@{}l@{}}2D\\ depth \\ map\end{tabular} & \begin{minipage}{0.13\textwidth}
      \includegraphics[width=\textwidth, height=20mm]{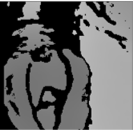}
    \end{minipage}
    &
    \begin{minipage}{0.13\textwidth}
      \includegraphics[width=\textwidth,height=20mm]{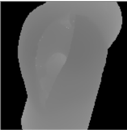}
    \end{minipage}
    & 
     \begin{minipage}{0.13\textwidth}
      \includegraphics[width=\textwidth, height=20mm]{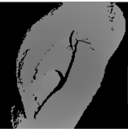}
    \end{minipage}
    \\ \hline
    \begin{tabular}[m]{@{}l@{}}3D\\ point \\ cloud\end{tabular} & \begin{minipage}{0.13\textwidth}
      \includegraphics[width=\textwidth, height=20mm]{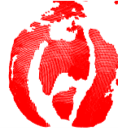}
    \end{minipage}
    &
    \begin{minipage}{0.13\textwidth}
      \includegraphics[width=\textwidth, height=20mm]{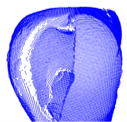}
    \end{minipage}
    & 
     \begin{minipage}{0.13\textwidth}
      \includegraphics[width=\textwidth, height=20mm]{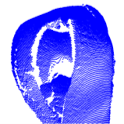}
    \end{minipage}
    \\ \bottomrule
  \end{tabular}
  \caption{Examples of simulated depth data represented in 2D depth map and 3D point cloud.}
\label{fig:dropout}
\end{table}

We then turned our attention to the RoI prediction network. Compared to the reference network, our network is different by having two mid-layer auxiliaries and a multi-box loss for training. We removed the mid-layer auxiliaries from the proposed architecture and trained the modified network on the proposed dataset. As shown in \autoref{tab:ablation}, the network can still learn from simulated images (\ie with an improvement from 32.95\% to 74.80\%), but the prediction accuracy was reduced by around 20\%. The degradation implies the importance of mid-layers for prediction accuracy, but not for synthetic-to-real transfer. Therefore, we speculate that the learning ability on synthetic data mainly comes from the training on multi-box loss. In fact, our RoINet localises the target in RGB frames by multi-box classification rather than direct regression. The classification is more tolerant of inconsistent features on different domains.

\section{Experiments on Markerless Tracking}
\subsection{Implementation}
While the network inference was scripted in Python, for faster speed, the BICP registration was coded in C++ and compiled into a dynamic linked library (DLL) that could be called in Python. 
Executed on a computer (IntelR\textcopyright CoreTMi5-8250U processor) with no dedicated graphics processing unit,
each RoI prediction took approximately 0.01s, the point segmentation took approximately 0.04s, and the BICP registration took approximately 0.05s.
Two threads were executed in parallel for the frame acquisition and inference, and the BICP registration, respectively.  
Given the RealSense camera's 30 Hz frame rate, the overall markerless tracking update frequency was found to be around 12 Hz. 

The same setup shown by the red path in \autoref{fig:hip} was used to obtain the gt femur pose for accuracy evaluation. The pre-scanned model $\mathbf{s}$ was first registered to the manually digitised bone surface for the initial pose $\mathbf{P}_{\scriptscriptstyle{gt}}^{\scriptscriptstyle{(A)}}$, then transformed into $\mathit{M_{\scriptscriptstyle{f}}}$ as a time-invariant local pose $\mathbf{P}_{\scriptscriptstyle{gt}}^{\scriptscriptstyle{(M_{\scriptscriptstyle{f}})}}$. The registered initial gt pose can be updated continuously based on optical tracking:
\begin{equation}
     \mathbf{P}_{\scriptscriptstyle{gt}}^{\scriptscriptstyle{(D)}}(t) 
    =  \prescript{\scriptscriptstyle D}{\scriptscriptstyle{M_{\scriptscriptstyle{d}}}}{\mathbf{T}}
    \times \prescript{\scriptscriptstyle {M_{\scriptscriptstyle{d}}}}{\scriptscriptstyle{A}}{\mathbf{T}}(t) 
    \times \prescript{\scriptscriptstyle A}{\scriptscriptstyle{M_{\scriptscriptstyle{f}}}}{\mathbf{T}}(t) 
    \times \mathbf{P}_{\scriptscriptstyle{gt}}^{\scriptscriptstyle{(M_{\scriptscriptstyle{f}})}}
\end{equation}
The real-time tracking error was defined as the relative transformation between the markerless-tracked femur pose and the gt pose in $D$:
\begin{equation}
\mathbf{P}_{\scriptscriptstyle{err}} = {\mathbf{P}_{\scriptscriptstyle{gt}}^{\scriptscriptstyle{(D)}}(t)}^{\scriptscriptstyle{-1}} \times \mathbf{P}^{\scriptscriptstyle{(D)}}(t)
\end{equation}
$\mathbf{P}_{\scriptscriptstyle{err}}$ was decomposed into the 3D rotational and translational misalignment. During each experiment, the RGB-D camera was held by a tripod and randomly placed at 10 different locations around the target knee. More than 50 frames of evaluated $\mathbf{P}_{\scriptscriptstyle{err}}$ were collected from each camera position to quantify the overall tracking error.

\subsection{Comparison with the Literature}
\begin{figure}[ht]
  \centering
\begin{subfigure}{.23\textwidth}
  \centering
  \includegraphics[width=\linewidth]{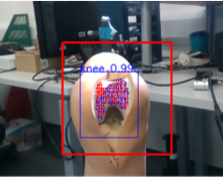}  
  \caption{Not occluded}
\end{subfigure}
\begin{subfigure}{.23\textwidth}
  \centering
  \includegraphics[width=\linewidth]{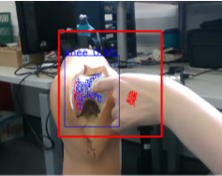}  
  \caption{Occluded by hand}
\end{subfigure}
\caption{Overlaid markerless segmentation (predicted RoI and segmented points) by Liu \etal  (red) and our networks (blue).}
\label{fig:scene}
\end{figure}

The RealSense D415 camera was first tested on the same model knee used for synthetic data creation. The markerless tracking proposed in \cite{heliu} was implemented and tested under the same setup, as a reference for performance comparison.
As shown in \autoref{fig:scene}, without target interaction, both reference and proposed methods track properly. When the femur is partially occluded by the hand, the RoI centre predicted by the reference RGB network drifts slightly from the actual femur centre. Fortunately, as the fixed cropping size (\ie 160) is large enough at the working distance, the cropped depth frames may still contain the target femur. However, the reference segmentation network fails to identify the femur pixels, resulting in an unreliable pose. In contrast, both the proposed RoI prediction and segmentation networks work well under hand occlusion. 

\begin{figure}[htbp]
\centering
\includegraphics[width =0.5\textwidth]{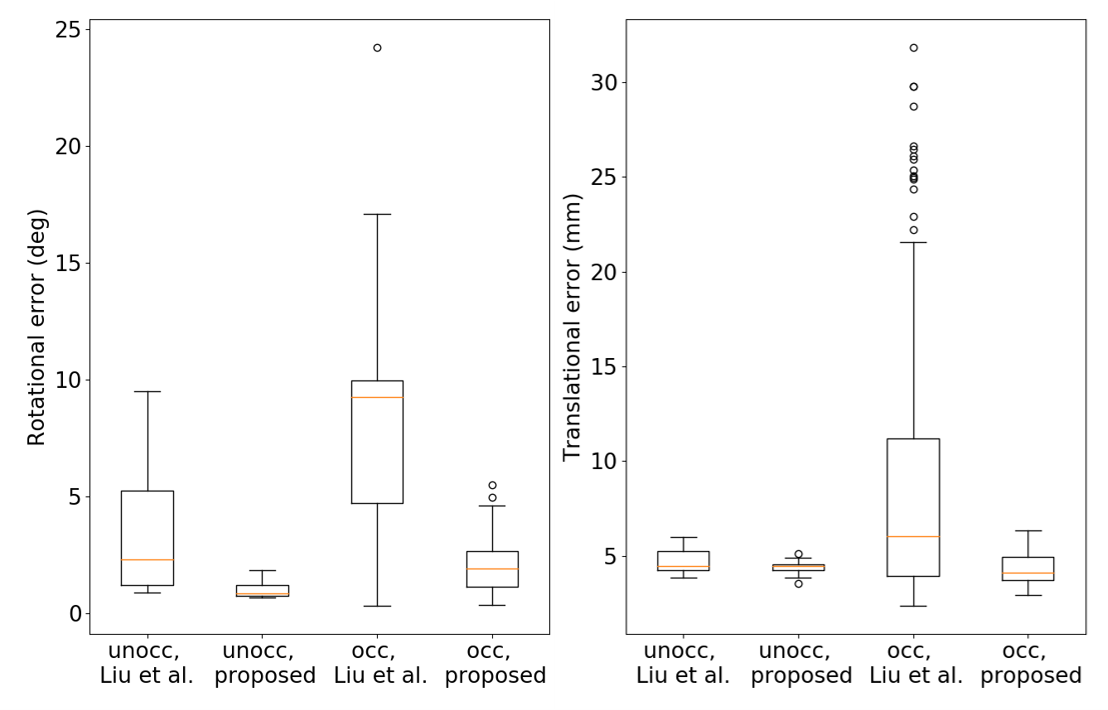}
\caption{Accuracy of our proposed method and the reference method by Liu \etal \cite{heliu} with/without target occlusion.}
\label{fig:all}
\end{figure}

\autoref{fig:all} compares the BICP-based markerless tracking accuracy obtained by the proposed and reference segmentation networks. The Kruskal-Wallis test was used to check whether the difference between obtained results is statistically significant. No matter whether the target occlusion exists, the proposed tracking can achieve better accuracy than the reference tracking (p-values$<$0.001 in both rotation and translation). The proposed markerless tracking achieves 1.02$^{\circ}\pm$0.33$^{\circ}$, 4.39 mm$\pm$0.33 mm error with no occlusion (unocc), and 2.05$^{\circ}\pm$1.10$^{\circ}$, 4.33 mm$\pm$0.78 mm error under occlusion (occ). There is no significant difference in translation (p-value = 0.21) but in rotation (p-value$<$0.001). 

\subsection{Camera Agnostic Performance}
Despite the promising results, the RealSense D415 camera is not designed for highly precise tasks. Therefore, a more accurate depth camera should be adopted for future clinical applications. To test the generalisability of the proposed network on new cameras, and to show the potential of markerless tracking in achieving higher accuracy, we deployed the trained network with an Acusense RGB-D camera (Revopoint 3D Technologies Inc.) that claims sub-millimetre accuracy within the 1-meter working distance. The Acusense camera, based on the coded IR structured light technology, has a higher RGB resolution (600$\times$800) and much longer focal length (\eg $f_x$=2061 compared to $f_x$=460 for RealSense camera). 
We subsampled the raw RGB frames into 300$\times$400 and padded the margin by white pixels into the designed input size of 360$\times$640 for RoINet. The predicted box corners were then mapped to the depth frames for cropping. Given the camera’s much lower frame rate of around 6-7 Hz, the overall markerless tracking reached a 5-6 Hz refresh rate with multi-threading computation. 

\begin{figure}[ht]
\begin{subfigure}{.23\textwidth}
  \centering
  \includegraphics[width=\linewidth]{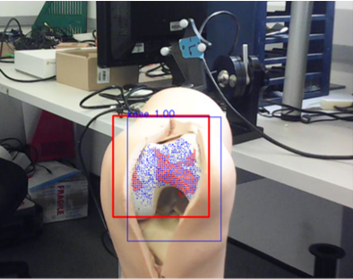}
  \caption{Not occluded}
\end{subfigure}
\begin{subfigure}{.23\textwidth}
  \centering
  \includegraphics[width=\linewidth]{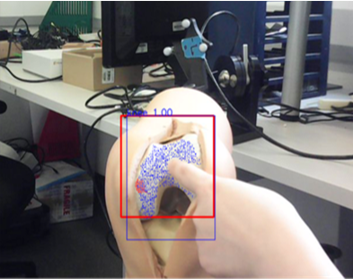}
  \caption{Occluded by hand}
\end{subfigure}
\caption{Overlaid markerless segmentation by Liu \etal (red) and proposed networks (blue) with a new Acusense camera.}
\label{fig:newcam}
\end{figure}

\begin{figure}[htbp]
\centering
\includegraphics[width =0.5\textwidth]{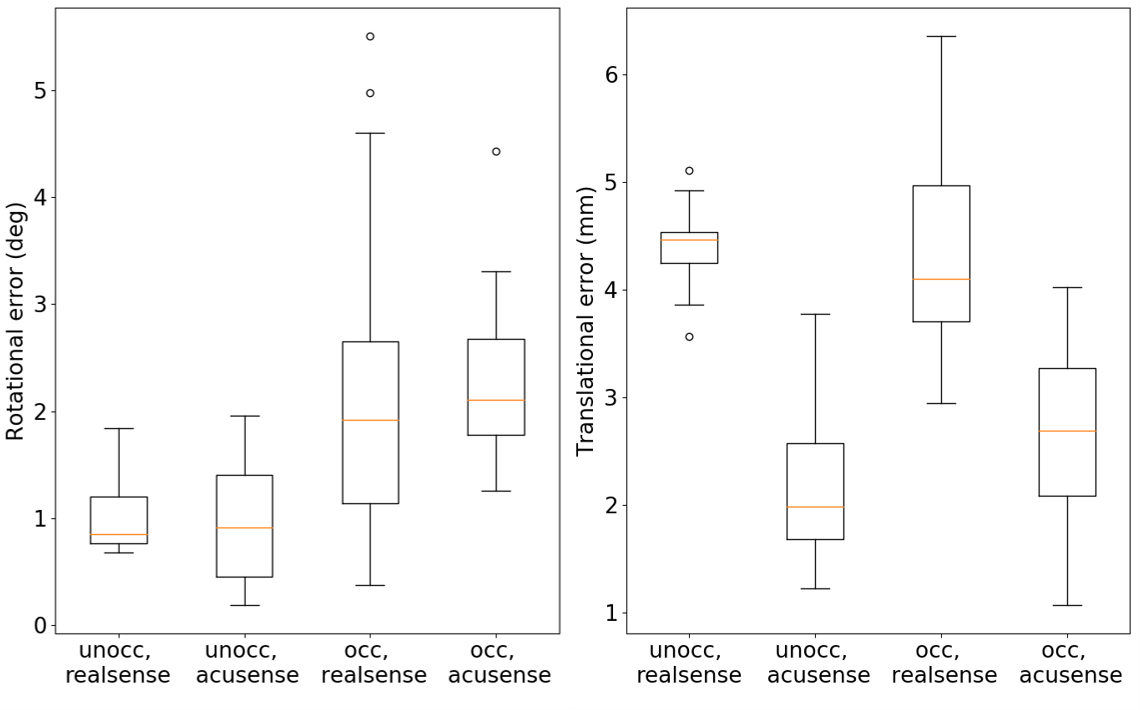}
\caption{Accuracy of the proposed markerless tracking with/without occlusion, tested on different cameras.}
\label{fig:newcambox}
\end{figure}

\autoref{fig:newcam} demonstrates the strength of our tracking over the proposed method by Liu \etal \cite{heliu} regarding device dependency or the lack thereof. While the fixed-size cropping by \cite{heliu} fails to cover the full target, our dynamic RoINet efficiently adapts to a larger cropping size. The segmentation network in \cite{heliu} is also less robust than our SegNet, which could be caused by the different features in 2D depth maps, since the Acusesne camera has a higher spatial resolution and less dropout effect around the edges. \autoref{fig:newcambox} shows how the tracking accuracy changes after using a more precise RGB-D camera. There is a significant accuracy improvement in translation (p-values$<$0.001 for both occ and unocc) but not in rotation (unocc: p-value = 0.25; occ: p-value = 0.24). The markerless tracking error is 0.95$^{\circ}\pm$0.55$^{\circ}$, 2.17 mm$\pm$0.62 mm with no occlusion, and 2.24$^{\circ}\pm$0.73$^{\circ}$, 2.62 mm$\pm$0.85 mm under occlusion. According to the quantitative score table for guide concepts proposed by Audenaert \etal \cite{audenaert2011custom}, the accuracy obtained here is in the clinically ``acceptable'' range (\ie error less than 4$^{\circ}$ and 4 mm).

\subsection{Generalisation Ability}
A general question is whether the network will still work if the target anatomy is different from the model/s used for training. We tested the qualitative segmentation performance on a cadaveric knee during a partial joint replacement dissection study (approved by Imperial College Healthcare NHS Trust Tissue Bank with the number R15022 for the use of human cadavers), where we had no ground truth to compare to, and quantitative tracking accuracy on a new (and different) model knee. Both of the targets had never been seen by the network during training. 
\begin{figure}[ht]
\begin{subfigure}{.23\textwidth}
  \centering
  \includegraphics[width=\linewidth]{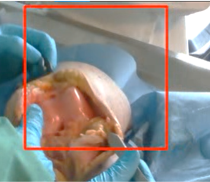}
  \caption{By Liu \etal}
\end{subfigure}
\begin{subfigure}{.23\textwidth}
  \centering
  \includegraphics[width=\linewidth]{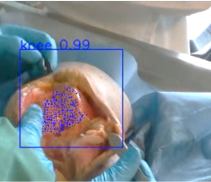}
  \caption{By ours}
\end{subfigure}
\caption{Markerless segmentation by Liu \etal (red) and proposed networks (blue) on a new cadaver knee under occlusion. Results are shown in pairwise recording.}
\label{fig:cadaver}
\end{figure}

\begin{figure}[htbp]
\centering
\includegraphics[width =0.5\textwidth]{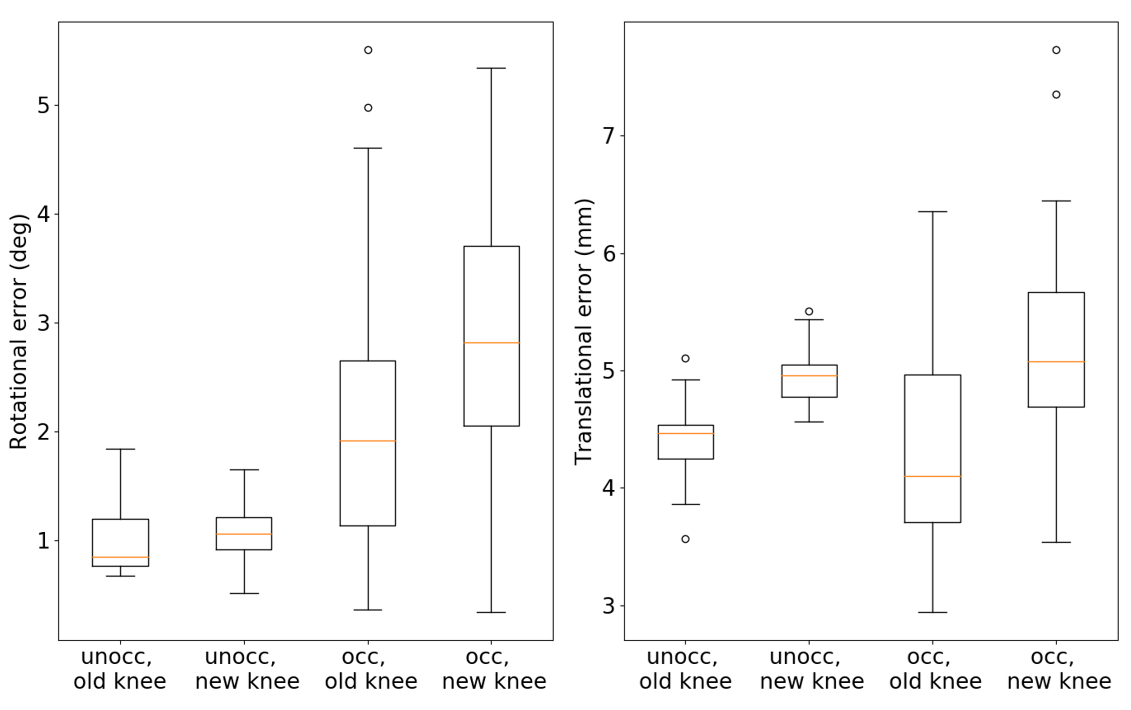}
\caption{Accuracy of the proposed markerless tracking with/without occlusion, tested on different model knees.}
\label{fig:newlegbox}
\end{figure}

 As shown in \autoref{fig:cadaver}, while the method proposed by Liu \etal fails under occlusion, 
our network gives reliable predictions. \autoref{fig:newlegbox} shows the quantified tracking accuracy by the proposed method on a new model knee. Although never seeing the target, the tracking accuracy remains high (i.e., 1.07$^{\circ}\pm$0.25$^{\circ}$, 4.94 mm$\pm$0.23 mm with no occlusion, and 2.82$^{\circ}\pm$1.22$^{\circ}$, 5.21 mm$\pm$0.83 mm with occlusion), indicating good generalisability to new geometry. Compared to the old knee, the new target experiences slightly higher tracking error in unoccluded translation, occluded rotation and occluded translation (p-values$<$0.05), suggesting that including more instances of target geometry for training may further improve the network performance.

\section{Conclusion and Future Work}
In this work, we proposed a new RGB-D based occlusion-robust markerless femur tracking method for computer-assisted knee surgeries. By training the network on a padded dataset with synthetic images, the robustness to target occlusion is learned in a cost and effort-efficient way. To ensure effective synthetic-to-real transfer, we show that the multi-box loss is critical for RoI prediction and learning on the 3D point cloud is vital for robust segmentation. While the state-of-the-art markerless tracking fails under target occlusion, our method can achieve a stable accuracy of around 2$^{\circ}$ and 4 mm no matter whether the target is fully visible to the camera or not. The proposed tracking can be deployed on new target geometries (including a cadaver knee) and with new RGB-D cameras without the need for network retraining. Consequently, we demonstrated here that, by simply using a more precise camera, we could achieve a tracking error of around 1-2$^{\circ}$ and 2-4 mm, a marked performance improvement that now meets the requirements for clinical deployment. The results indicate the possible use of RGB-D imaging as a new modality for surgical applications.

Our synthetic training data can be further improved for better network performance. The imported knee model is currently considered as a rigid body with a fixed femur exposure. By modelling the skin part as a non-rigid body controlled by some critical nodes, various extents of skin exposure could be included in the synthetic images. Besides, as suggested by results on a new knee, including more target geometries in simulation could enrich the generated data. Finally, we will fully demonstrate the overall markerless navigation workflow in a cadaveric study in the future.

\bibliographystyle{ieeetr}
\bibliography{ref}

% \end{IEEEbiography}



\end{document}